\documentclass{article}

\PassOptionsToPackage{numbers}{natbib}

\usepackage[preprint]{neurips_2024}

\usepackage[utf8]{inputenc} 
\usepackage[T1]{fontenc}    
\usepackage{hyperref}       
\usepackage{url}            
\usepackage{booktabs}       
\usepackage{amsfonts}       
\usepackage{nicefrac}       
\usepackage{microtype}      
\usepackage{xcolor}         

\usepackage{amsmath}
\usepackage{bbm}

\usepackage[ruled,vlined,linesnumbered]{algorithm2e}

\usepackage{algpseudocode}

\usepackage{graphicx}

\DeclareMathOperator*{\argmax}{arg\,max}

\title{Array-Based Monte Carlo Tree Search}

\author{%
  James Ragan\\
  California Institute of Technology\\
  Pasadena, CA 91125 \\
  \texttt{jragan@alumni.caltech.edu} \\
  \And
  Fred Y. Hadaegh \\
  California Institute of Technology\\
  Pasadena, CA 91125 \\
  \texttt{hadaegh@caltech.edu} \\
  \And
  Soon-Jo Chung \\
  California Institute of Technology\\
  Pasadena, CA 91125 \\
  \texttt{sjchung@caltech.edu} \\
}

\begin{document}

\maketitle

\begin{abstract}
  Monte Carlo Tree Search is a popular method for solving decision making problems. Faster implementations allow for more simulations within the same wall clock time, directly improving search performance. To this end, we present an alternative array-based implementation of the classic Upper Confidence bounds applied to Trees algorithm. Our method preserves the logic of the original algorithm, but eliminates the need for branch prediction, enabling faster performance on pipelined processors, and up to a factor of 2.8 times better scaling with search depth in our numerical simulations.
\end{abstract}

\section{Introduction}

Monte Carlo Tree Search (MCTS), provides efficient, anytime, and asymptotic solutions to sequential decision making problems formulated as Markov Decision Processes (MDPs). It has been applied to a broad range of planning problems~\cite{Browne_2012,swiechowski2023monte} from super human performance in games~\cite{silver2016mastering}, to real time planning in dynamical systems~\cite{doi:10.1126/scirobotics.ado1010} to even to predicting protein structures~\cite{bryant2022predicting}. MCTS works by constructing a tree of simulated possible futures while biasing the tree towards actions leading to high rewards~\cite{Browne_2012,swiechowski2023monte}. In doing so, the root node's value estimate converges to the true optimal value, allowing for computationally tractable and iteratively better approximate solutions, even for problems where closed form solutions do not exists or are infeasible to compute. 

The classical Upper Confidence bounds applied to Trees (UCT) result provides value convergence at $\mathcal{O}(\log(N)/N)$~\cite{kocsis2006bandit}, while more recent results for Fixed-Depth MCTS provides value convergence at a polynomial rate $\mathcal{O}(N^{-1/2})$~\cite{Shah_2020}. Therefore, increasing the number of simulations performed within a computation budget will result in improved value estimates and performance. However, changes to the underlying algorithm will change these theoretical results. 

Instead, we present an alternative array-based implementation designed to optimize the branching performance of MCTS on pipelined processors. A key motivation of our work is the difficulty of deploying MCTS on hardware accelerators such as GPUs due to high memory latency and limited branching performance~\cite{pharr2005gpu}, with software packages often providing limited branching capability at all~\cite{jax2018github}.

Branches are a type of control flow commonly found in programs where the execution path taken depends (or branches) on the result of a previous instruction. One of the most common occurrences of branching is an "if" statement, which splits execution based on a boolean value. Due to limitations of pipelined architectures, which are nearly ubiquitous on modern processors, branches result in decreased computational performance as visualized in Figure~\ref{fig:Processor Pipelining} below. To leverage the full performance of a CPU (or GPU), instructions (such as adding two numbers) are queued and processed in order through different sections of the processor simultaneously. When a branch occurs, the processor must guess which execution path will be followed next, and load the appropriate instructions. If the guess is incorrect, these instructions must be flushed, and new instructions loaded at the start of the pipeline, resulting in lost computation cycles~\cite{10.1145/356683.356687}, a problem further exasperated by memory latency~\cite{4362179}. As a result, despite using the same logic, different algorithm implementations can have significantly different computational performance when deployed to real systems, leading to the development of techniques to avoid expensive instructions~\cite{anderson2005bit,han2011reducing,10.1145/3570638}. 

\begin{figure}
    \centering
    \includegraphics[width=0.75\linewidth]{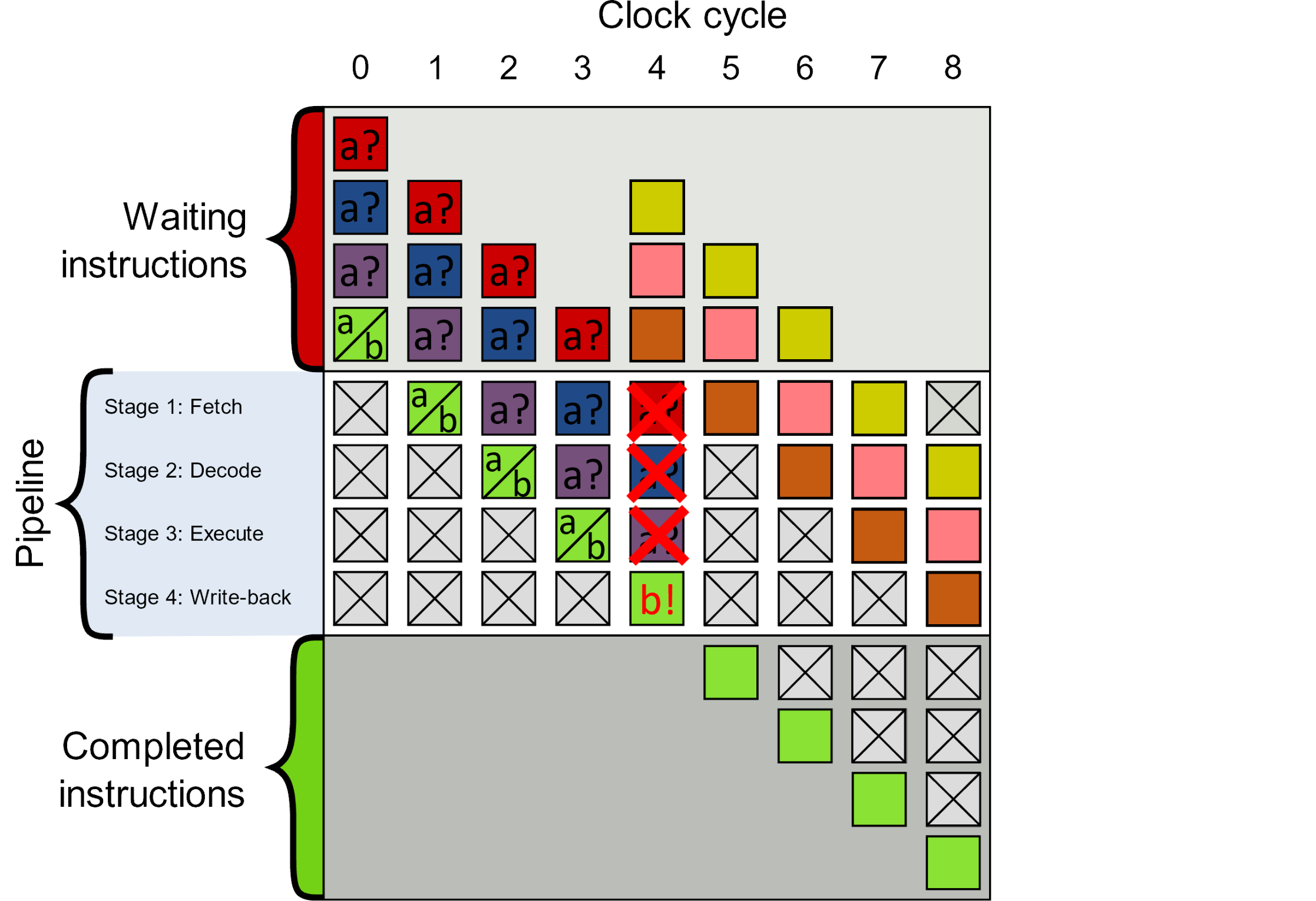}
    \caption[Processor Pipelining.]{\textbf{Processor Pipelining.} A pipelined processor carries out instructions (represented as colored blocks) in stages. Here, each column represents a processor clock cycle. Between clock cycles, instructions are advanced through the pipeline, allowing each component of the hardware to be utilized simultaneously. When branching between multiple execution paths occurs, the processor guesses which instructions to load. If it guesses incorrectly, the instructions must be flushed and replaced with the correct execution path. \footnotemark }
    \label{fig:Processor Pipelining}
\end{figure}

Branching is fundamental to tree searches, which decide which child node to proceed to (or new child to initialize), at each level of the tree, during every simulation. Our alternative implementation of UCT avoids the need for branch prediction, mitigating the performance impact. Our contributions are summarized as follows:
\begin{enumerate}
    \item We present algorithms for predictable child selection in MCTS.
    \item We use these algorithms to create an array-based implementation of MCTS which does not require branch prediction
    \item We validate our method through numerical experiments and show we improve scaling with simulation depth by up to a factor of 2.8 over an equivalent tree-based implementation.
\end{enumerate}
\footnotetext{Adapted from "A generic 4-stage pipeline" by Cburnett under \url{https://creativecommons.org/licenses/by-sa/3.0/}}

\section{Related work}

\paragraph{Parallelization} A common method to accelerate tree search is through parallelization to take advantage of multi-threaded computing algorithms and processors. 
The challenge however, is that tree search methods are fundamentally sequential, as each search uses information gained during the previous backpropagation step. The simplest approach is root parallelization~\cite{chaslot2008parallel}, where multiple independent tree searches are run. On completion, the value functions at each root node are averaged to determine the overall optimal value~\cite{cazenave:hal-02310186}. While conceptually simple, this approach is inefficient, as it results in many repeated computations. Leaf parallelization instead parallelizes roll outs from newly added leaf nodes across multiple threads~\cite{chaslot2008parallel,cazenave:hal-02310186}. This minimizes repeat computation, but can only be employed at the end of the tree. Other approaches perform multiple searches within the same tree, but must carefully manage the communication between threads to avoid data corruption, introducing overhead~\cite{chaslot2008parallel}. Still other approaches combine several of these techniques, such as by using CPUs for in-tree parallelization, and GPUs for leaf parallelization~\cite{6009083,6932879,cai2021hyp}. 

\paragraph{Tree search heuristics} Other approaches seek out heuristics to guide the tree search for more efficient exploration. Learned value functions were used by Alpha Go to approximate the outcome of a game from an intermediate position, giving MCTS earlier signals to bias its exploration towards~\cite{silver2016mastering}. Similar approaches have been taken in real-time motion planning, where an expert tree search is used to train a neural network enabling more efficient online planning~\cite{riviere2021neural}. Other methods leverage domain specific knowledge to modify the tree search, by limiting actions, setting sub goals, or imposing constraints~\cite{swiechowski2023monte}. 

While they achieve substantial empirical performance increases, both parallelization and heuristic methods modify the underlying search algorithm, and therefore loose the theoretical guarantees of traditional MCTS search~\cite{kocsis2006bandit} or more recent analysis~\cite{Shah_2020}. This inspires our approach of optimizing the implementation of MCTS algorithms instead. As we do not need to modify the search algorithm, we can run the original search algorithm and retain the theoretical guarantees or apply our implementation to these parallelization and heuristic methods as well. 

 \paragraph{Arrays in tree searches} Arrays have previously been proposed to compress tree searches in the context of string matching for natural language processing~\cite{31365}. Branchless search algorithms have also previously been identified as desirable in the context of binary search problems~\cite{10.1007/978-3-642-38527-8_13}, to avoid the penalties of branch misprediction on pipelined architectures~\cite{10.1145/356683.356687}. However, these methods have yet to be applied to MCTS or decision making problems.

\section{Preliminaries}

\begin{figure}
    \centering
    \includegraphics[width=0.6\linewidth]{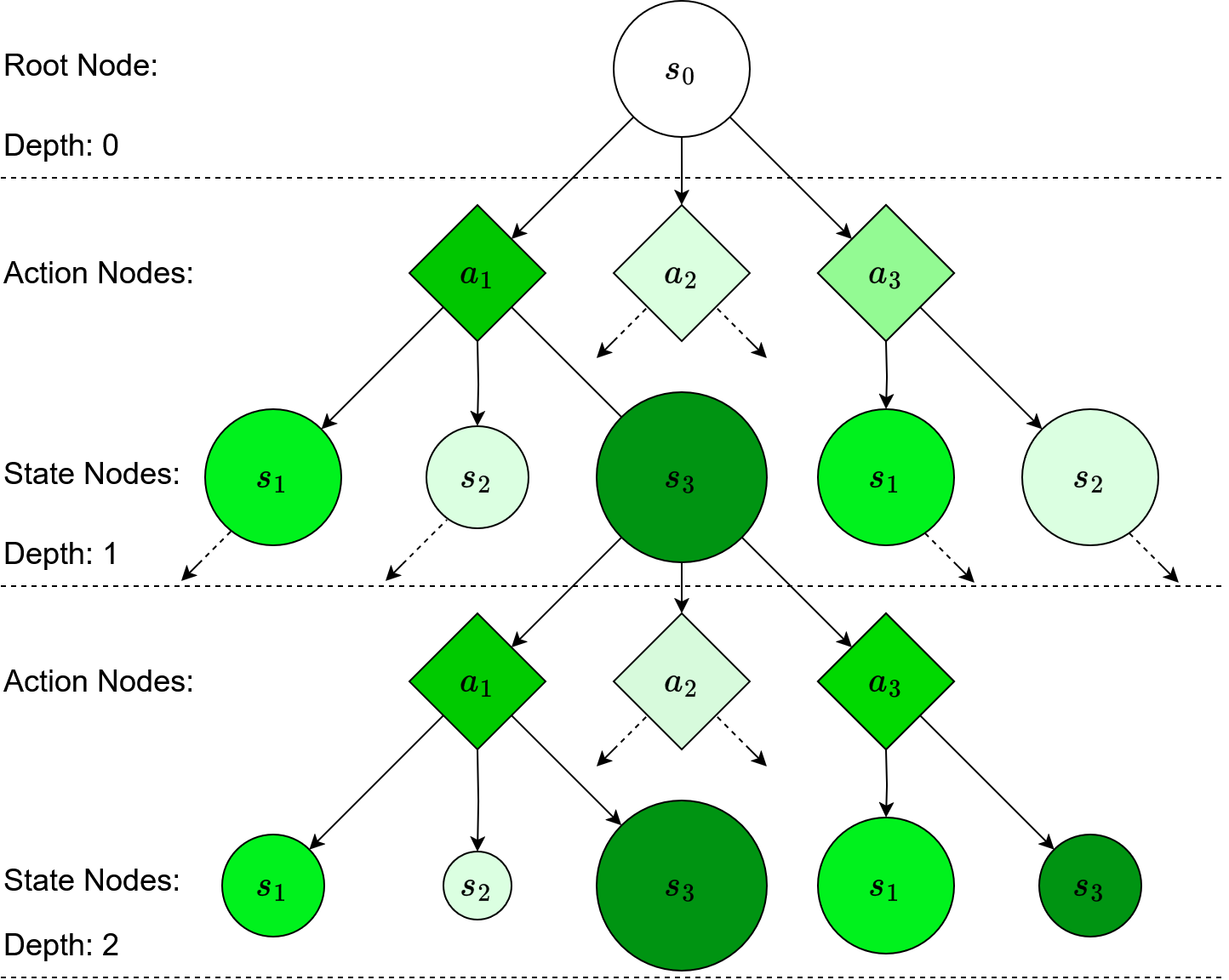}
    \caption[MDP tree search diagram.]{\textbf{MDP tree search diagram.} Diagram of a tree search over a simple MDP with 3 actions and 3 states. The tree has been split into depth layers, with depth 0 containing only the root node. Node size represents relative visit frequency. Darker colored states represent higher rewards, and darker action nodes represent higher values resulting from the weighted average of the rewards encountered under them. Note that each action may not result in all possible states.}
    \label{fig:MDP diagram}
\end{figure}

MCTS over a MDP consists of action nodes and state nodes, which together make up a single layer of the tree, as shown in Fig.~\ref{fig:MDP diagram} for a simple system with three actions and three states. Each node stores the total number of visits so far, the value or reward at the node, and the node's children. To simulate down the tree, an action is selected according to the UCT augmented value function:

\begin{equation}
    \hat{V}_\textup{aug}(H \cup a) = \hat{V}(H \cup a) + c \sqrt{\frac{\log N(H)}{N(H \cup a)}}
    \label{eq:UCT value}
\end{equation}

where $H$ represents the history in the tree up until the current state node, and $H \cup a$ represents taking action $a$ at this point in the tree. $\hat{V}(\cdot)$ and $N(\cdot)$ represent the value estimates and visit counts for given histories, respectively. The next state is generated by simulating the system forward using the selected action and compared to all the current child state nodes. If no match is found, a new child is initialized. The search continuous until the simulation depth has been reached, then backpropagates up the tree, adjusting the value and visit counts as it goes. Implementing this algorithm with a tree data structure provides a natural method for branching over possible child nodes. However, this branching is hard to predict.

\section{Algorithm Overview}

Our objective is to construct a version of MCTS with a predictable execution path by using arrays as the primary data structure. This will allow for better processor performance, by avoiding branch prediction failures in the processor pipeline and by leveraging existing optimizations for array operations. As a motivating example, Algorithm~\ref{algo:branchlessinspiration} demonstrates using mathematical operations to avoid the need for an "if" statement. This form of micro optimization is valuable on processors which feature expensive branching, have limited access to compiler optimizations, or lack conditional move instructions~\cite{branchlessWithBitMaskExample}.

\begin{algorithm}
\label{algo:branchlessinspiration}
\newcommand{\intE}{\KwSty{int8}}
\newcommand{\bool}{\KwSty{bool}}
\intE{} \texttt{a}~;~\intE{} \texttt{b}~;~\bool{} \texttt{c}~;~\intE{} \texttt{r} \;
\If{\texttt{c}}{
 \texttt{r} = \texttt{a}\;
}\Else{
 \texttt{r}= \texttt{b}\;
}
\tcc{Equivalent Implementation}

\texttt{r} = \texttt{a} * \texttt{c} + \texttt{b} * !\texttt{c} \;
\tcc{Equivalent Implementation}

\intE{} \texttt{m} = \texttt{c} \;

\texttt{m}= \texttt{m} | \texttt{m} << 1\;
\texttt{m} = \texttt{m} | \texttt{m} << 2\;

\texttt{m} = \texttt{m} | \texttt{m} << 4\;

\texttt{r} = \texttt{a} \& \texttt{m}| \texttt{b} \& \texttt{m} \; 
\caption{Three equivalent implementations of selecting the integer \texttt{a} or \texttt{b} based on the value of the condition \texttt{c}. The first implementation introduces a branch, the last two do not.}
\end{algorithm}

In MCTS, the fundamental instance of branching is deciding whether to add a child node or not. If the selected action (or state), has already been added to the tree, then the previous node is retrieved and the algorithm traverses downwards. Otherwise, a new node must be initialized before continuing. Unfortunately, this occurrence of branching exhibits different behavior, as opposed to simply resulting in different values like in Algorithm~\ref{algo:branchlessinspiration}.

\begin{figure}
    \centering
    \includegraphics[width=0.4\linewidth]{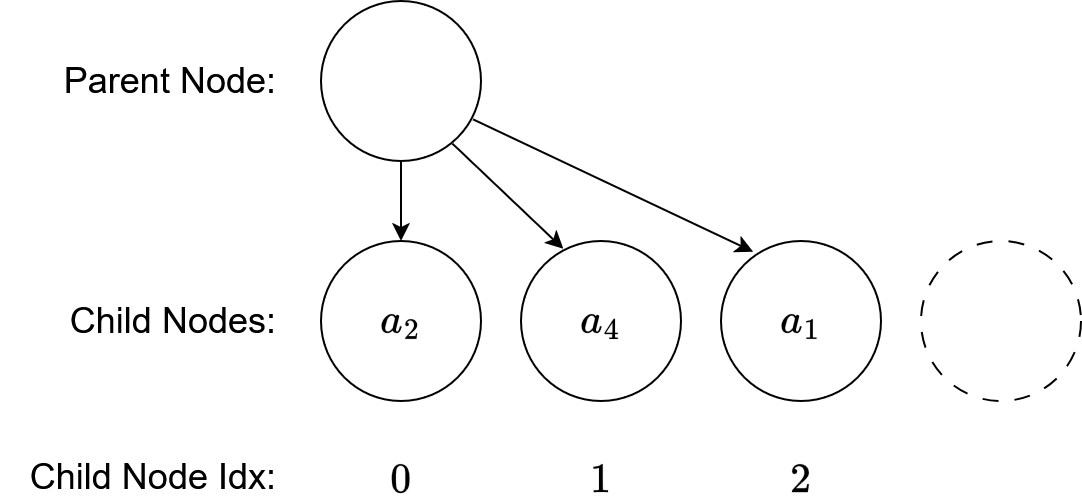}
    \caption[Child node diagram.]{\textbf{Child node diagram.} Child nodes can be thought of as an indexed list. In this case, taking action $a_4$ would be equivalent to selecting index 1, whereas taking action $a_0$ would be represented by index 3, indicating that child has not yet been added to the list.}
    \label{fig:child diagram}
\end{figure}

However, this example still gives a clue to how the branching can be made predictable. If we considered the child nodes to be arranged in a list ordered by time of insertion as diagrammed Fig.~\ref{fig:child diagram}, then matching an existing node is equivalent to finding the index of the matching child within this list, whereas the index past the end of the list represents a new node to initialize. Following the example in Algorithm~\ref{algo:branchlessinspiration}, the index of the matching child is \texttt{a}, \texttt{b} is the new node, and \texttt{c} is now whether or not a match occurred. Note that while the match index, \texttt{a}, may be ill defined when no match exists, there is no issue because its value will be discarded.

When adding a new node to the tree, we may need to perform some initialization steps, or otherwise execute different behavior than when selecting an existing node. It is still possible to avoid guessing which execution path to follow, but it now comes at a price. Instead of performing one set of behavior conditioned on whether the node is a new or existing child, we perform both behaviors every time, and use techniques like that of Algorithm~\ref{algo:branchlessinspiration} to nullify the operations that should not be performed.

\section{Array-Based MCTS Algorithm Details}

\begin{figure}
    \centering
    \includegraphics[width=0.6\linewidth]{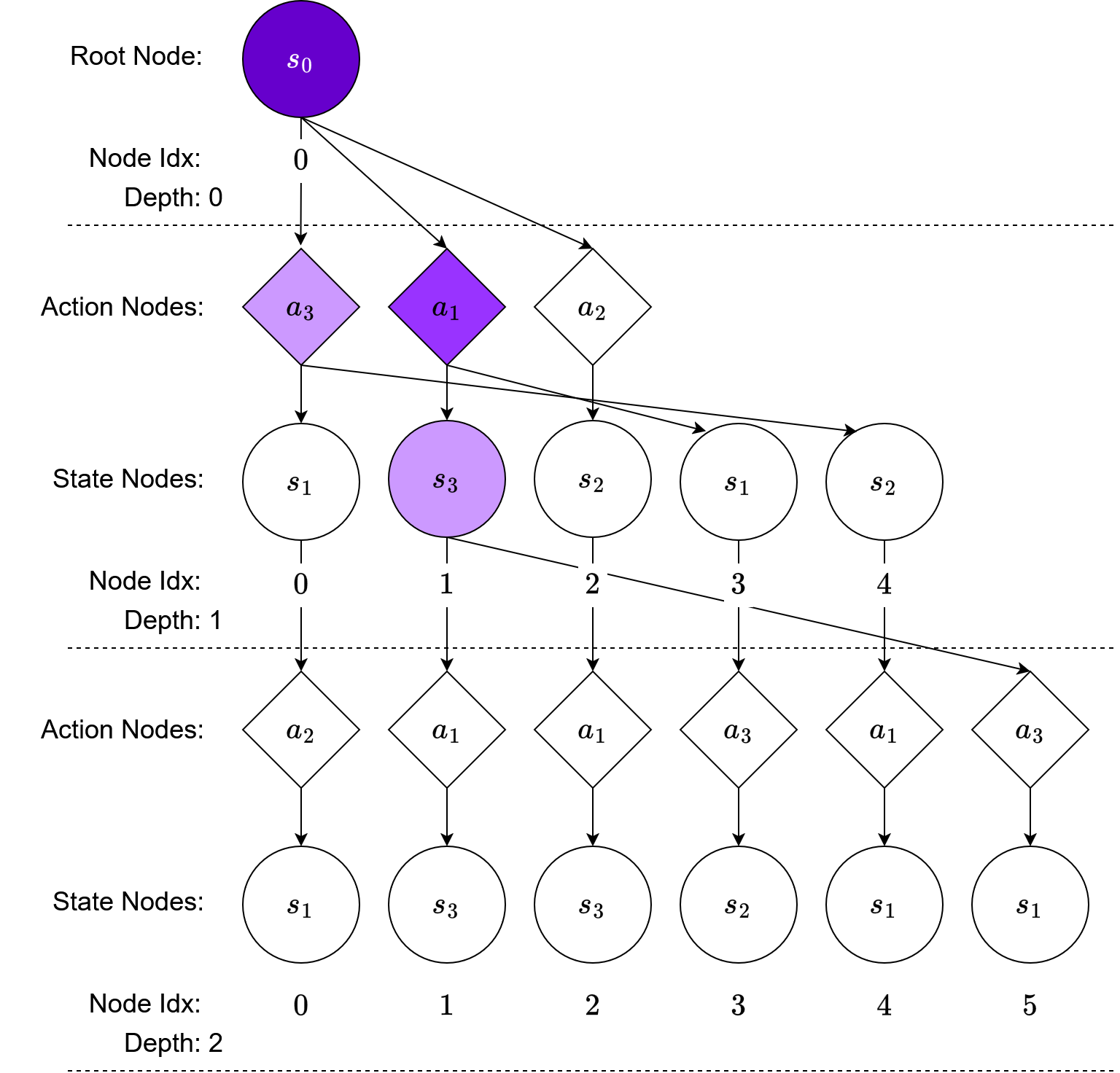}
    \caption[MDP node array diagram.]{\textbf{MDP node array diagram.} Diagram of the first six simulations of the same tree search as shown in Fig.~\ref{fig:MDP diagram}. Each node is now sorted by type within a layer and assigned an index when added. Darker colors represent the visit frequency of each node. The values and rewards are not visualized here.}
    \label{fig:Array MDP diagram}
\end{figure}

Our algorithm, organizes the nodes by layer within the tree search, separating action and state nodes into two arrays. Now, adding a node to the tree means adding a new node to the end of the appropriate array, and parent nodes save the index of each child. This modified structure is visualized in Fig~\ref{fig:Array MDP diagram} for the first six iterations of a search over the simple three action and three state system we considered earlier. The nodes in each layer are now ordered by time of creation, with the children of different parents being intermixed. 

Note that the length of each node array increases by layer. At depth 1, the action nodes are fully expanded, as all three actions which can be taken from the root node have been tried. Conversely, four more state nodes can be added to the first layer to reach a total of nine, three for each of the three actions. These nine state nodes can themselves have three children, and so on. This observation that the branching is bounded, if growing geometrically by layer, motivates our hierarchical structure where the data from each layer is stored in separate arrays, as opposed to storing all state and all action nodes within the same array. The smaller layers near the root node can be stored efficiently, increasing the likelihood they are kept within the processor cache, leading to lower latency and faster execution. 

\subsection{Action Branching}

To store the action node children of a state node, we take advantage of the fact that the set of available actions is constant and numbered throughout the tree search. To do so, we consider a vector of child indexes for each individual state node, resulting in a 2D array for each layer. For a given entry, the value represents the location of the child action node in the next layer, the column indicates the parent state node, while the row index indicates the action taken. All entries are are initialized to a flag value, indicating that action has not yet been tried. 

Our action selection method is shown in Algorithm~\ref{algo:actionSelection} below. Given the current depth and state index, we retrieve the indexes of each child action (the corresponding action node column). If unvisited actions exist (indicated by the flag value), \texttt{untriedAction} is true and we sample from the untried actions. If none exist, we retry the action with the best UCT value.

\begin{algorithm}[h!]
\label{algo:actionSelection}
\SetKwProg{Def}{def}{:}{}
\newcommand{\intE}{\KwSty{int8}}
\newcommand{\bool}{\KwSty{bool}}
\Def{selectChildAction(\texttt{depth},\texttt{curStateIdx})}{
\texttt{childActionIdxs} = \texttt{childActionNodes}(\texttt{depth}).column(\texttt{curStateIdx}) \;
\texttt{childActionValues} = \texttt{actionValues}(\texttt{depth} + 1)[\texttt{childActionIdxs}] \;
\texttt{childActionVisits} = \texttt{actionVisits}(\texttt{depth} + 1)[\texttt{childActionIdxs}] \;

\bool{} \texttt{untriedAction} = anyUnvistedChild(childActionVisits) \;
\texttt{uctValues} = uctValue(\texttt{childActionValues},

~~\texttt{childActionVisits},\texttt{stateVisits}(\texttt{depth})[\texttt{curStateIdx}]) \;
\intE{} \texttt{bestAction} = maxIndex(\texttt{uctValues}) \;
\intE{} \texttt{newAction} = getRandomUntriedAction(\texttt{childActionVisits}) \;

\intE{} \texttt{nextAction} = \texttt{newAction} * \texttt{untriedAction}  + \texttt{bestAction} * !\texttt{untriedAction} \;
\intE{} \texttt{nextActionIdx} = \texttt{numActionsAtDepth}(\texttt{depth} + 1) * \texttt{untriedAction}  + \texttt{childActionNodes}(\texttt{depth})[\texttt{bestAction},\texttt{curStateIdx}] * !\texttt{untriedAction} \;

\texttt{childActionNodes}(\texttt{depth})[\texttt{nextAction},\texttt{curStateIdx}]  =  \texttt{nextActionIdx} \;

\texttt{numActionsAtDepth}(\texttt{depth} + 1)  +=  \texttt{untriedAction} \;
\textbf{return} \texttt{nextAction} \;
}

\caption{Predictable Action Selection.}
\end{algorithm}

In the above algorithm, \texttt{depth} and \texttt{curStateIdx} represent the current position of the tree search, \texttt{actionValues}, \texttt{actionVisits}, \texttt{stateVisits}, \texttt{numActionsAtDepth}, and 
\texttt{childActionNodes} represent global tree data arrays we can access by layer, and we use the helper functions anyUnvistedChild, uctValue, and getRandomUntriedAction to determine if unvisited child actions exist, compute the UCT value function (Eq.~\eqref{eq:UCT value}) for each action, and return a random untried action (or a default value if none exist), respectively. Note we only need to return the selected \texttt{nextAction}, as all other information has been stored in the global tree data arrays.

\subsection{State Branching}

We use a similar method for the child state nodes. However, this is complicated by the fact that the states are not already ordered like the actions are. To address this, we make two modifications to the  array representing the child state nodes from the action node architecture. First, while each column still corresponds to a given parent node, the rows no longer have a consistent meaning. Instead state nodes are added to each column in order of generation. To track the number of state nodes already added to each column, we add a row to the bottom of the array which counts the number of state nodes already present, initialized to zero. This doubles as the index which a new state should be added to.

The other change we make from the child action node array is to assume a max level of state branching. In simple MDPs, way may know this branching exactly. However, in continuous dynamical systems (particularly with unbounded noise) the branching may be unbounded~\cite{sFEAST}. 
For a given number of simulations and noise level, however, we can conservatively estimate a maximum level of branching and verify it empirically. Further, we can also set different levels of assumed branching by search layer, knowing that actions closer to the root will be taken more often, leading to more branching 
Other approaches could be taken to bound the amount of state branching, including clipping the noise distribution or varying the amount of noise or discretization in different layers of the tree. 

The resulting child state node selection algorithm is given in Algorithm~\ref{algo:stateSelection} below. 
Given the depth, action index, and a multi-deminsional state generated by simulation, we begin by retrieving the first $N_{S,l}$ (the assumed state branching at layer $l$) child state indexes from the column corresponding to the current action node. Any unassigned states are initialized to point to the last possible state in the layer, which has a value initialized to NaN. This ensures these entries will not match when we compare each child state against the generated state. We set the \texttt{matchIdx} to the child state matching the most dimensions of the given state, however the \texttt{matchFlag} is only set if all dimensions match. As with the action selection, we use this flag to distinguish between whether to write to \texttt{matchIdx} or a new entry in the child state node array, as well as determine the value of \texttt{nextStateIdx} and if we should update the number of state nodes in the layer and the current column of the child state node array. 

\begin{algorithm}[H]
\label{algo:stateSelection}
\SetKwProg{Def}{def}{:}{}
\newcommand{\intE}{\KwSty{int8}}
\newcommand{\bool}{\KwSty{bool}}
\Def{selectChildState(\texttt{depth},\texttt{curActionIdx},\texttt{generatedState})}{
\texttt{childStateIdxs} = \texttt{childStateNodes}(\texttt{depth}).column(\texttt{curActionIdx}) \;
\texttt{childStates} = \texttt{stateNodes}(\texttt{depth})[\texttt{childStateIdxs}] \;
\For{$i=1 \hdots N_{S,l}$}{
       \texttt{stateMatches}[i] = $\Sigma$ (\texttt{childStates}[i] == \texttt{generatedState}) \;
}

\intE{} \texttt{matchIdx} = maxIndex(\texttt{stateMatches}) \;

\bool{} \texttt{matchFlag} = \texttt{stateMatches}[\texttt{matchIdx}] == size(\texttt{generatedState})\;

\intE{} \texttt{idxInChildArray} = \texttt{matchIdx} * \texttt{matchFlag}  + \texttt{childStateNodes}(\texttt{depth})[ $N_{S,l}$,\texttt{curActionIdx}] * !\texttt{untriedAction} \;

\intE{} \texttt{nextStateIdx} = \texttt{childStateNodes}(\texttt{depth})[\texttt{matchIdx},\texttt{curStateIdx}] * \texttt{matchFlag}  + \texttt{numStatesAtDepth}(\texttt{depth}) * !\texttt{matchFlag} \;

\texttt{childStateNodes}(\texttt{depth})[\texttt{nextStateIdx},\texttt{curActionIdx}]  =  \texttt{nextStateIdx} \;
\texttt{stateNodes}(\texttt{depth})[\texttt{nextStateIdx}] = \texttt{generatedState} \;

\texttt{childStateNodes}(\texttt{depth})[$N_{S,l}$,\texttt{curActionIdx}]  +=  !\texttt{matchFlag}  \;

\texttt{numStatesAtDepth}(\texttt{depth})  +=  !\texttt{matchFlag} \;
\textbf{return} \texttt{nextStateIdx} \;
}
\caption{Predictable Child State Selection.}
\end{algorithm}

In the above algorithm, \texttt{curStateIdx} is the index of the action node we are deteriming the child of, \texttt{generatedState} is our generated state to compare against, \texttt{stateNodes}, \texttt{numStatesAtDepth}, and \texttt{childStateNodes} represent global tree data arrays we can access by layer, $\Sigma$ sums the number of matched states, maxIndex() returns the index of the best matching child state, and size() returns the number of elements in the state. There is no issue with non-unique maxima for \texttt{stateMatches} as this only occurs when no exact matches exist. Again we only need to return the selected \texttt{nextStateIdx}, all other information has been stored in the global tree data arrays.

\subsection{Array-Based MCTS Implementation}

Combining Algorithms~\ref{algo:actionSelection} and~\ref{algo:stateSelection} results in our full array-based MCTS algorithm, visualized below in Fig.~\ref{fig:Full Array MDP diagram}. Each layer consists of 8 arrays, divided between the action and state nodes, with columns representing a single action or state within a layer. The parent state and action arrays store each node's parent index. Alternatives to Algorithms~\ref{algo:actionSelection} and~\ref{algo:stateSelection} can be constructed by searching over the parent arrays, however these require exhaustive searches that scale poorly with the number of nodes in a layer, and return varying numbers of nodes, which could reintroduce the need for branch prediction. 

\begin{figure}
    \centering
    \includegraphics[width=.8\linewidth]{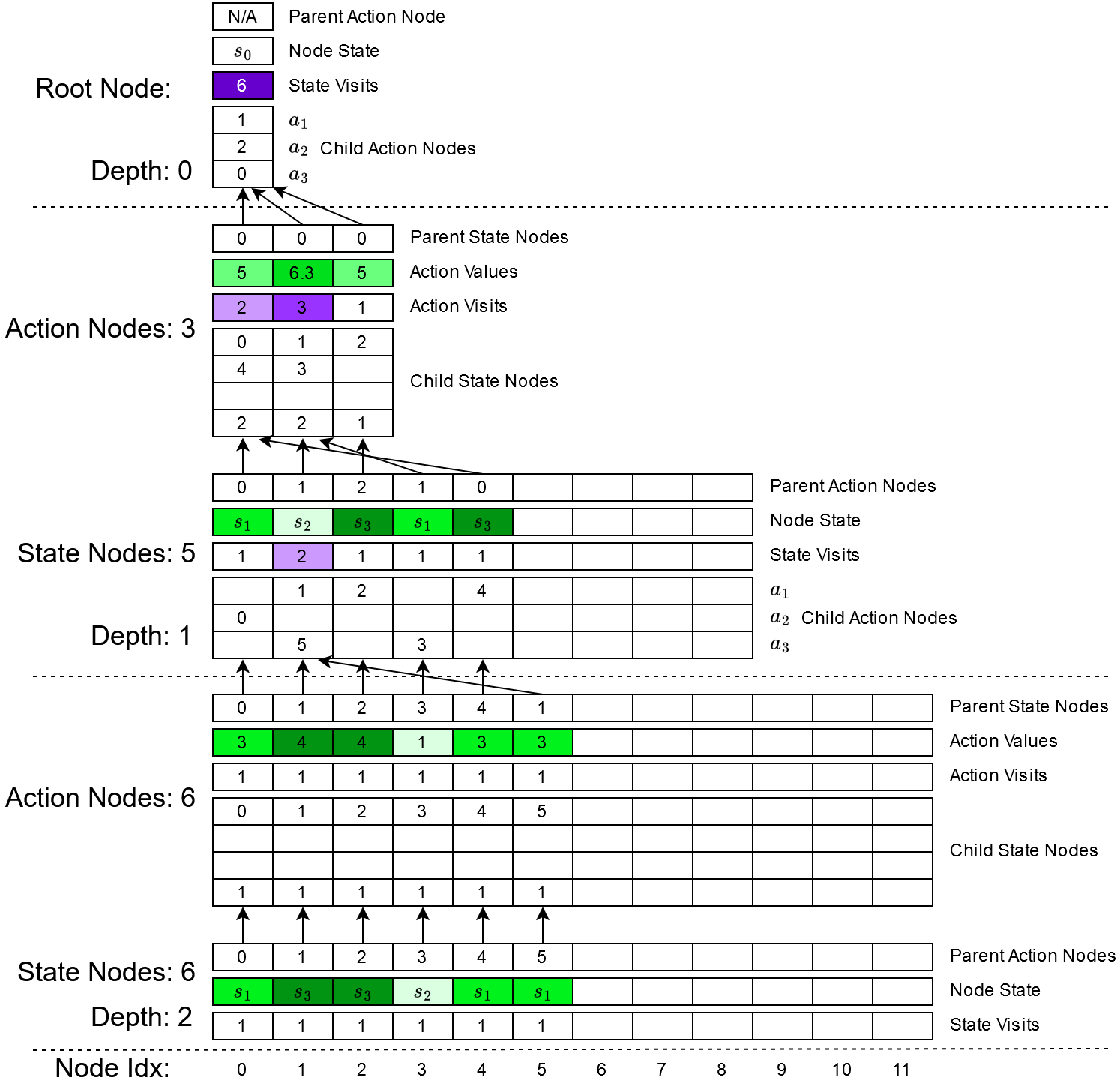}
    \caption[Array-Based MCTS Diagram.]{\textbf{Array-Based MCTS Diagram.} Diagram of the first six simulations of of the same tree search as shown in Figs.~\ref{fig:MDP diagram} and~\ref{fig:Array MDP diagram}, now using an array-based MCTS representation. Each layer consists of 8 arrays, 4 each corresponding to action and state nodes. Each column corresponds to a state or action node within the layer. The parent arrays give the indexes of each node's parent shown by arrows. The child action arrays store the indexes of the child actions in the layer below for each available action, if any. The row of the child index indicates the action taken (note the actions are 1-indexed). The child state arrays point to the child state nodes that have resulted from a given action. The states nodes are sorted by order of creation, with the last element of each column indicating what index the next child should be inserted at. The columns in layers 0 and 1 are limited by the maximum possible branching due to the number of actions and states. The number of columns in layer 2 is limited for visualization. For clarity, array entries which have not been written to during the search are left blank.  Darker colored states represent higher rewards, and darker action values represent higher weighted averages of the rewards encountered after the taken action. Darker visit counts represent a higher frequency of visiting each node. The final layer has no child actions.}
    \label{fig:Full Array MDP diagram}
\end{figure}

We present the complete pseudocode of our array-based MCTS algorithm below in Algorithm~\ref{algo:arrayBasedMCTS}. We first initialize the data arrays for each layer of the search to the default values of zero, unless stated otherwise. We then iterated down the tree, using Algorithms~\ref{algo:actionSelection} and~\ref{algo:stateSelection} to select the action and state child nodes before backpropagating up the tree in the standard MCTS fashion. This process is repeated the specific number of simulations, and the action with the best value is then returned.

\begin{algorithm}
\label{algo:arrayBasedMCTS}
\SetKwProg{Def}{def}{:}{}
\newcommand{\intE}{\KwSty{int8}}
\newcommand{\bool}{\KwSty{bool}}
\Def{arrayBasedMCTS(\texttt{rootState},$N$,\texttt{maxDepth},\texttt{actionSet},$N_{S,1-\texttt{maxDepth}}$)}{
\tcc{Initializing the Root Node}

\texttt{stateNodes}(0) = \texttt{rootState}\;
\texttt{childActionNodes}(0) = const(size(\texttt{actionSet}),1,size(\texttt{actionSet})) \;
\tcc{Initializing Tree Data Arrays at Each Depth}

\texttt{curMaxChildNodes} = size(\texttt{actionSet})\;
\For{$l = 1 \hdots \texttt{maxDepth}$}{
    \texttt{curMaxNodes} = \texttt{curMaxChildNodes}\;
    \texttt{curMaxChildNodes} = min($N$,\texttt{curMaxChildNodes}*$N_{S,l}$)\;
    
    \texttt{childStateNodes}($l$) = const($N_{S,l}+1$,\texttt{curMaxNodes},\texttt{curMaxChildNodes}-1) \;
    \texttt{childStateNodes}($l$)[$N_{S,l}$,:] = 0\;
    \texttt{curMaxNodes} = \texttt{curMaxChildNodes}\;
    \texttt{stateNodes}($l$)[\texttt{curMaxNodes}-1] = \texttt{NaN} \;
    \texttt{curMaxChildNodes} = min($N$,\texttt{curMaxChildNodes}*size(\texttt{actionSet}))\;
    
    \texttt{childActionNodes}($l$) = const(size(\texttt{actionSet}),\texttt{curMaxNodes},\texttt{curMaxChildNodes}) \;
}
\For{$i = 1 \hdots N$}{
    \texttt{curStateIdx} = 0 \;
    \texttt{curState} = \texttt{rootState} \;
    \tcc{Simulate Down the Tree}
    
    \For{$l = 0 \hdots \texttt{maxDepth}-1$}{
        \texttt{curAction} = $selectChildAction$($l$,\texttt{curStateIdx}) \;
        \texttt{curActionIdx} = \texttt{childActionNodes}($l$)[\texttt{curAction},\texttt{curStateIdx}]\;
        \texttt{stateParentNodes}($l + 1$)[\texttt{curActionIdx}] = \texttt{curStateIdx}\;
        \texttt{curState} = simulate(\texttt{curState},\texttt{actionSet}[\texttt{curAction}])\;
        \texttt{curStateIdx} = $selectChildState$($l + 1$,\texttt{curActionIdx},\texttt{curState})\;
        \texttt{actionParentNodes}($l + 1$)[\texttt{curStateIdx}] = \texttt{curActionIdx}\;
    }
    \texttt{summedReward} = 0\;
    \tcc{Backpropagate Up the Tree}
    
    \For{$l = \texttt{maxDepth}-1 \hdots 0$}{
        \texttt{summedReward} += rewardFunction(\texttt{stateNodes}($l + 1$)[\texttt{curStateIdx}]\;
        \texttt{curActionIdx} = \texttt{actionParentNodes}($l + 1$)[\texttt{curStateIdx}]\;
        \texttt{actionVisits}($l + 1$)[curActionIdx] += 1\;
        \texttt{actionValues}($l + 1$)[curActionIdx] += (\texttt{summedReward} -\texttt{actionValues}($l + 1$)[curActionIdx] )/ \texttt{actionVisits}($l + 1$)[curActionIdx] \;

        \texttt{curStateIdx} = \texttt{stateParentNodes}($l + 1$)[\texttt{curActionIdx}]\;
        \texttt{stateVisits}($l$)[curStateIdx] += 1
    }
}
\textbf{return} $\argmax \texttt{actionValues}(0)$ \;
}
\caption{Array-Based MCTS}
\end{algorithm}

In Algorithm~\ref{algo:arrayBasedMCTS}, \texttt{rootState} is the initial state, $N$ the number of simulations to perform, \texttt{maxDepth} indicates how many layers are present in the tree, \texttt{actionSet} is the fixed set of actions for all layers, $N_{S,1-\texttt{maxDepth}}$ represents the state branching limits at each depth,  const($m$,$n$,$c$) is a function which creates an $m$ by $n$ matrix with each entry set to $c$, and rewardFunction() gives a reward based on the current state. The variables \texttt{curMaxNodes} and \texttt{curMaxChildNodes} capture the maximum possible branching at the current layer and its children. Knowing that we start with a single root node, there are at most size(\texttt{actionSet}) actions below it, and size(\texttt{actionSet})$*N_{S,1}$ state nodes below that, and so on, seen visually in Fig.~\ref{fig:Full Array MDP diagram}. We also know that at most one action and one state node can be added per layer, per simulation, leading to an alternative bound on the number of nodes per layer. Finally, we note that there is no discounting in the backpropagation step in Algorithm~\ref{algo:arrayBasedMCTS}, as we consider the finite horizon setting, but it can easily be added in.

\section{Empirical Results}

In this section, we validate our array-based implementation of MCTS by solving a continuous MDP. We compare our performance against the baseline of a traditional tree-based implementation, as well as an alternative array-based method which does not separate nodes by layers. For consistency, each search method was implemented in C++17, with the same compiler optimizations and libraries. All experiments were performed on a AMD Ryzen 7 3700X (8-Core Processor, 3593 Mhz) with 32 GB of memory available.

\begin{figure}
    \centering
    \includegraphics[width=.9\linewidth]{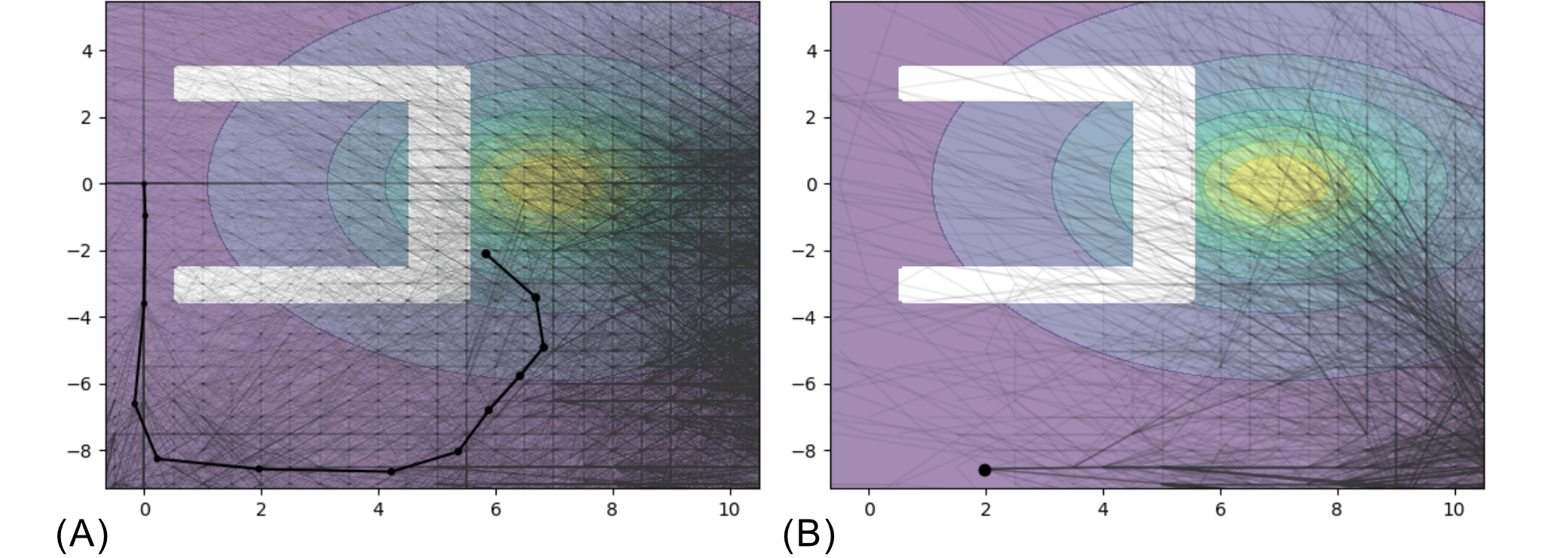}
    \caption[Array-based MCTS solving a non-convex problem.]{\textbf{Array-based MCTS solving a non-convex problem.} Tree search around a non-convex obstacle. (\textbf{A}) The array-based MCTS method successfully escapes the local minimum and navigates to the goal state. (\textbf{B}) One iteration of the array-based MCTS search. Note the state discretization results in branches snapping to the same points in space.}
    \label{fig:bug trap}
\end{figure}

We consider the challenging problem of navigating a three degree of freedom vehicle adapted from our previous work. A further challenge is that tree search must be able to consider far enough into the future to discover that going around the obstacle is optimal, as opposed to a simple gradient-based solution which will get stuck against the obstacle wall. In Fig.~\ref{fig:bug trap}(A), we present a 12 time step simulation, demonstrating our array-based method can successfully solve this MDP problem. In Fig.~\ref{fig:bug trap}(B) we show the growth of the array-based search over a single time step, validating that a branching search is being performed. 
Not that while some simulations result in points inside the obstacle, these receive a large negative penalty, and are not selected by the algorithm.

\begin{figure}
    \centering
    \includegraphics[width=.8\linewidth]{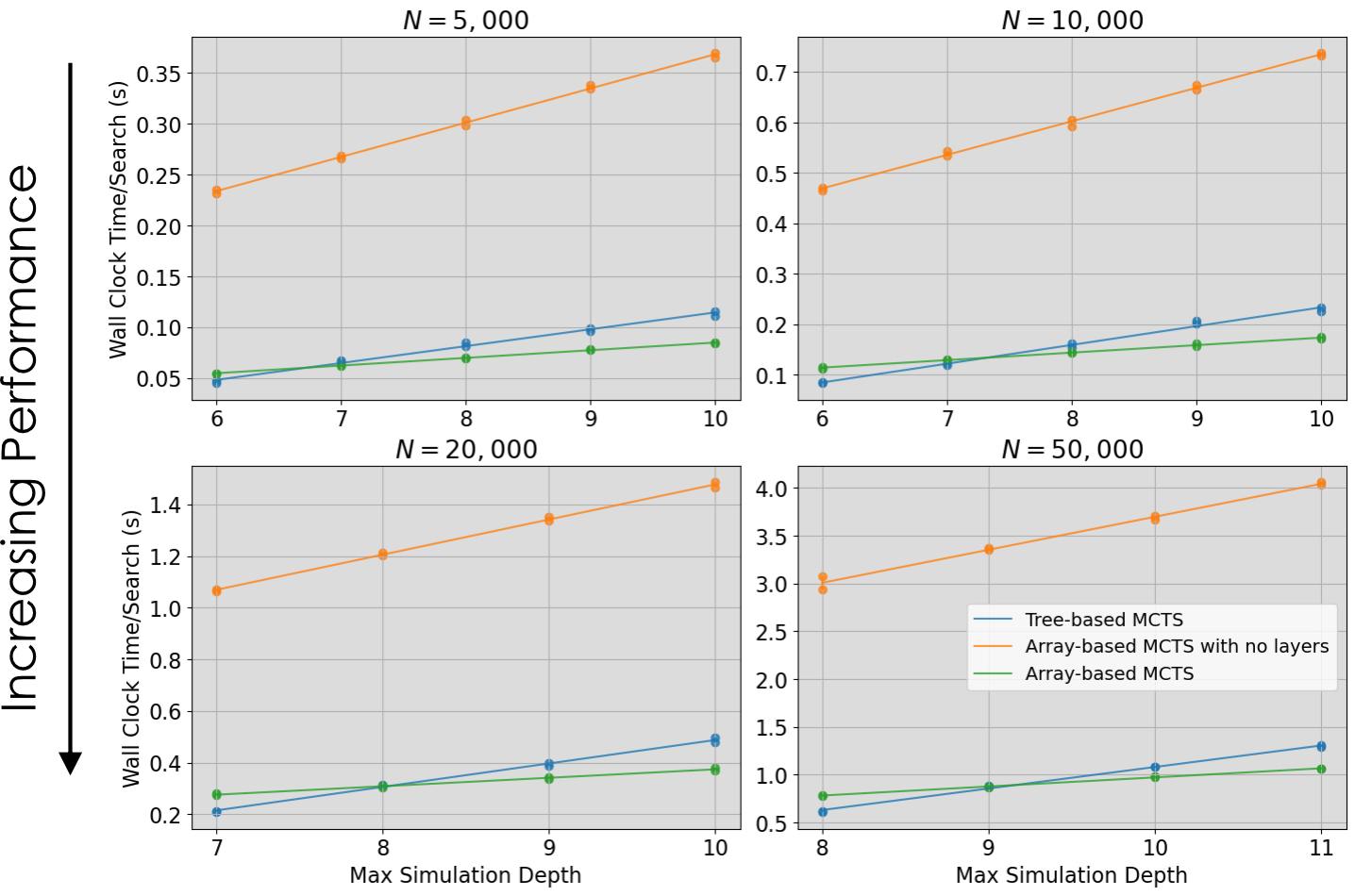}
    \caption[Array-based MCTS performance on the non-convex problem.]{\textbf{Array-based MCTS performance on the non-convex problem.} The wall clock time of our array-based MCTS search compared with a tree-based implementation, and an alternative array-based MCTS algorithm that does not sort its nodes into layers, presented over a range of simulation depths, $N$. All times are averaged over 100 time steps (10 simulations of 10 steps each). A lower wall clock time indicates better simulation performance.}
    \label{fig:MDP WCT}
\end{figure}

Next, we consider the wall-clock time performance of our array-based tree search on this MDP problem. In Fig.~\ref{fig:MDP WCT}, we present the results of averaging the simulation times of the first 10 time steps of this experiment over 10 trials. We note that across a wide range of total simulations, our method demonstrates consistently better scaling with increased simulation depth than the tree-based method as well as the array-based method with no layer sorting. Comparing linear fits between the tree-based method and the array-based method, we see a slope of 0.017 seconds per layer of the search when $N=5,000$ and 0.225 seconds per layer when $N=50,000$ for the tree-based method, compared with 0.008 and 0.095 seconds per layer for the array-based method. Taking the ratios of each methods' scaling with search depth, we see the array-based method outperforms the tree-based method by a factor ranging from 2.2 to 2.8 across the wide range of total simulations. 

Interestingly, the array-based method with no sorting by layer has the same or worse scaling as the tree-based implementation. Combined with its dramatically worse wall clock time, this suggest that sorting the nodes by layer is necessary for efficient memory access, and may indicate that the most frequently accessed nodes near the root of the search are in fact remaining within the processor cache.

Another general trend observed was that the tree-based search took longer during the later time steps of the experiment, while the array-based searches did not. This may be due to the tree more efficiently storing broader searches with more branching, which typically occurred at the beginning of the experiment, compared to the array-based implementations, which may be faster when the same values are repeatedly accessed in the CPU cache. 

\section{Conclusion}

We have introduced an array-based implementation of MCTS, and demonstrated that it out scales traditional tree-based implementations in challenging continuous MDP settings, while maintaining theoretical correctness and the associated convergence guarantees. 
Our results present several opportunities for future work. 

First, while recomputation is inevitable if we wish to avoid branch prediction, one way to minimize the redundant effort is to identify which layers will be fully expanded by the end of the tree search. A hybrid approach could first expand these layers completely before computing the rest of the tree. On subsequent simulations, these first layers could be efficiently bypassed, and our predictable array-based search applied only to the layers which will not be fully explored. 

Next, while GPUs and other hardware accelerators with limited branching capabilities were a motivating application for the development of our method, the results presented here are all run on a CPU. Deploying our method on hardware accelerators is the most obvious opportunity for future work, and we anticipate significant performance improvements can be achieved. Finally, dedicated profiling could be performed to identify other opportunities for algorithmic optimization, particularly to verify our intuition that keeping the portions of the tree close to the root in the processor cache is essential for rapid searches.

\begin{ack}

We thank Sina Aghli who shared an algorithm to avoid branching which inspired our predictable branching techniques.

\end{ack}

\medskip

{
\small

\setcitestyle{numbers}
\bibliographystyle{unsrtnat}
\bibliography{refs}
}

\end{document}